\documentclass[letterpaper]{article} % DO NOT CHANGE THIS
\usepackage{aaai23}  % DO NOT 
\usepackage{times}  % DO NOT CHANGE THIS
\usepackage{helvet}  % DO NOT CHANGE THIS
\usepackage{courier}  % DO NOT CHANGE THIS
\usepackage{graphicx} % DO NOT CHANGE THIS
\usepackage{natbib}  % DO NOT CHANGE THIS AND DO NOT ADD ANY OPTIONS TO IT
\usepackage{caption} % DO NOT CHANGE THIS AND DO NOT ADD ANY OPTIONS TO IT
\usepackage{booktabs}

\usepackage{microtype}
\usepackage{booktabs} % for professional tables
\usepackage{paralist}
\usepackage{calligra}
\usepackage[T1]{fontenc}
\usepackage{enumitem}

\usepackage{sidecap}
\usepackage{float}
\usepackage{amsmath}
\usepackage{amsthm}
\usepackage{amssymb}
\usepackage{bm}
\usepackage{algorithm}
\usepackage{algorithmic}
\usepackage{subcaption}
\usepackage{color}
\definecolor{airforceblue}{rgb}{0.36, 0.54, 0.66}
\definecolor{cinnamon}{rgb}{0.82, 0.41, 0.12}

\definecolor{orange}{rgb}{1.0, 0.5, 0.0}

\usepackage{algorithm}
\usepackage{algorithmic}

\usepackage{newfloat}
\usepackage{listings}
\DeclareCaptionStyle{ruled}{labelfont=normalfont,labelsep=colon,strut=off} % DO NOT CHANGE THIS
\floatstyle{ruled}
\newfloat{listing}{tb}{lst}{}
\floatname{listing}{Listing}
\title{Pruning the Unlabeled Data to Improve Semi-Supervised Learning}% for Deep Learning}
\author {
    Guy Hacohen,\textsuperscript{\rm 1,2}
    Daphna Weinshall\textsuperscript{\rm 1}
}
\affiliations {
    \textsuperscript{\rm 1} {School of Computer Science and Engineering, The Hebrew University of Jerusalem, Israel}\\
    \textsuperscript{\rm 2} {Edmond \& Lily Safra Center for Brain Sciences, The Hebrew University of Jerusalem, Israel}\\
    \{guy.hacohen,daphna@mail.huji.ac.il\}
}

\usepackage{bibentry}
\begin{document}

\maketitle

\begin{abstract}

In the domain of semi-supervised learning (SSL), the conventional approach involves training a learner with a limited amount of labeled data alongside a substantial volume of unlabeled data, both drawn from the same underlying distribution. However, for deep learning models, this standard practice may not yield optimal results. In this research, we propose an alternative perspective, suggesting that distributions that are more readily separable could offer superior benefits to the learner as compared to the original distribution. To achieve this, we present \emph{PruneSSL}, a practical technique for selectively removing examples from the original unlabeled dataset to enhance its separability. We present an empirical study, showing that although \emph{PruneSSL} reduces the quantity of available training data for the learner, it significantly improves the performance of various competitive SSL algorithms, thereby achieving state-of-the-art results across several image classification tasks.

\end{abstract}

\section{Introduction}
In recent years, extensive research has centered around the domain of deep semi-supervised learning (SSL), showcasing remarkable effectiveness across various domains. This success largely arises from the abundance of unlabeled data, in contrast with labeled data whose collection usually demands costly human annotation. Accordingly, most of the previous works focused on the creation and improvement of optimization algorithms that can utilize both labeled and unlabeled data. Differently, we focus on improving SSL performance by manipulating the unlabeled data directly.

Traditionally, SSL paradigms assume that both labeled and unlabeled data stem from the same underlying distribution. This is a reasonable assumption -- when there is a significant disparity between the unlabeled and labeled data, the learning model may learn the wrong dependencies across these datasets. Consequently, this misalignment can lead to errors in generalization, ultimately mitigating the advantages that could otherwise be gained from incorporating unlabeled data.

Our investigation indicates that reliance on this assumption may not be optimal in the context of SSL. In particular, we demonstrate that modifying the distribution of unlabeled examples to enhance their distinctiveness boosts the performances of numerous deep SSL algorithms. With this objective, we present \emph{PruneSSL}, a systematic approach designed to identify and prune unlabeled instances that undermine separability within the unlabeled dataset, thereby boosting overall performance.

More specifically, instances are considered to undermine the separability of the dataset if, given some meaningful embedding of the dataset and its corresponding labeling function, simple classifiers fail to accurately classify these instances when trained on the complete dataset.  Clearly, executing this process directly on the unlabeled dataset is unfeasible due to the absence of labels. Consequently, the challenge lies in devising a technique capable of identifying these instances without relying on a labeling function.

\begin{figure*}[htb!]
    \begin{subfigure}{.825\textwidth}
      \centering
      \includegraphics[width=1\linewidth]{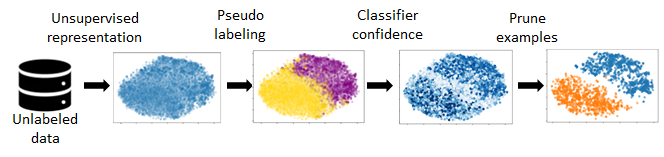}\vspace{-0.5cm}
      \caption{PruneSSL}
      % \label{subfig:clustering_matters_regular}
    \end{subfigure}
    \begin{subfigure}{.165\textwidth}
      \centering
      \includegraphics[width=1\linewidth]{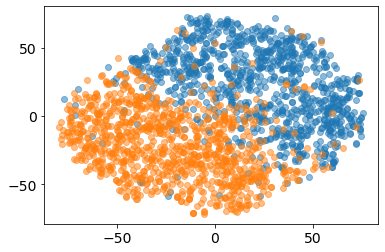}
      \caption{Random sample}
      % \label{subfig:clustering_matters_regular}
    \end{subfigure}    
    \caption{Flowchart of \emph{PruneSSL}. In (a), we provide a step-by-step visual representation of how \emph{PruneSSL} operates on the binary CIFAR-10 dataset. To aid visualization, each step is projected onto a 2-dimensional space using t-SNE. \emph{PruneSSL} starts with an unsupervised representation task, followed by pseudo-labeling denoted in yellow and purple. Subsequently, a simple classifier is trained on the data, assigning confidence levels to each example -- darker shades of blue indicate higher confidence. Finally, examples with low confidence are pruned from the data, with the orange-blue color scheme corresponding to actual labels. In contrast, (b) displays pruning using a random function. Evidently, \emph{PruneSSL} results in a distribution that is more separable.
    }
    \label{fig:tsne_visualization}
    \vspace{0.2cm}
\end{figure*}

Our proposed method, called \emph{PruneSSL}, starts by employing a deep representation task over the unlabeled data, resulting in a meaningful embedding space. Subsequently, it calculates pseudo-labels for the unlabeled data using a deep-clustering algorithm or k-means within the embedding space. Equipped with both the embedding space and the pseudo labels, \emph{PruneSSL} trains a simple classifier on the data and prunes instances that elicit the lowest levels of confidence from the classifier.

To illustrate this concept, we present a step-by-step visualization of each stage of \emph{PruneSSL} in Fig.~\ref{fig:tsne_visualization}, utilizing a binary subset extracted from CIFAR-10. The figure shows a comparison between random instance pruning from the unlabeled dataset and the \emph{PruneSSL} approach. For visualization purposes, the data is projected into a 2-dimensional space using t-SNE. Note that this projection serves exclusively for visualization purposes, as all computations occur within the original embedding space. Evidently, even in the projected space, the resulting data exhibits increased separation, as the 2 clusters in this case are more distinct. The exact details of this experiment are described in Section~\ref{sec:results}. 

The inherent modularity of \emph{PruneSSL} presents clear advantages as well as potential drawbacks. On the positive side, this modularity facilitates seamless integration of upcoming developments in both self-representation and pseudo-labeling tasks. Moreover, the framework readily adapts to diverse domain-specific challenges by tailoring the representation task accordingly. However, a notable drawback of \emph{PruneSSL} lies in its reliance on the effectiveness of both the self-representation task and the pseudo-labeling process.

In Section~\ref{sec:results} we present the outcome of our empirical investigation, demonstrating the effectiveness of \emph{PruneSSL}. Despite its reduction of the unlabeled dataset's size, we find that a diverse array of SSL algorithms gain significant performance boosts when trained with the pruned unlabeled set, compared to training with the entire unlabeled dataset. Notably, these advantages are even more pronounced when replacing the pseudo-labeling process with an oracle possessing knowledge of the actual unlabeled dataset labels. These results demonstrate the validity of the idea behind \emph{PruneSSL}.

In Section~\ref{sec:ablation_study}, we describe the results of an ablation study, designed to assess the significance of each component within the \emph{PruneSSL} algorithm, as well as the varying influence of several hyperparameters. Our findings reveal a consistent trend: \emph{PruneSSL} consistently produces comparable qualitative outcomes across a diverse range of embedding spaces and pseudo-labeling techniques. Additionally, our analysis highlights an interesting discovery –- while an optimal value exists for the number of instances to be pruned, this parameter displays robustness, enabling the use of an extensive array of values with comparable effectiveness. Finally, we report that the benefits of \emph{PruneSSL} increase as the size of the labeled set decreases.

\begin{figure*}[htb!]
    \begin{subfigure}{.32\textwidth}
      \centering
      \includegraphics[width=1\linewidth]{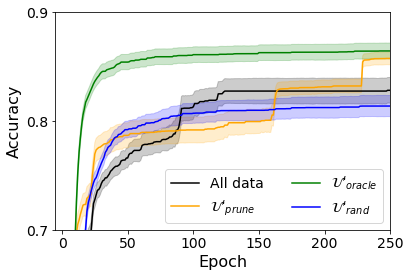}
      \caption{CIFAR-10}
      % \label{subfig:clustering_matters_regular}
    \end{subfigure}
    \begin{subfigure}{.32\textwidth}
      \centering
      \includegraphics[width=1\linewidth]{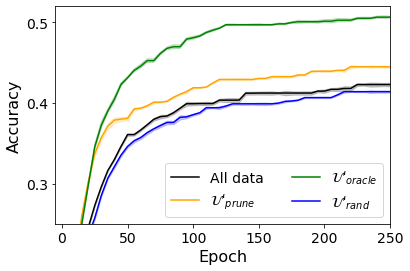}
      \caption{CIFAR-100}
      % \label{subfig:clustering_matters_regular}
    \end{subfigure}
    \begin{subfigure}{.32\textwidth}
      \centering
      \includegraphics[width=1\linewidth]{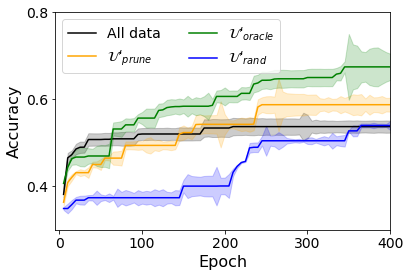}
      \caption{STL-10}
      % \label{subfig:clustering_matters_inverse}
    \end{subfigure}
    \caption{
    Comparison of \emph{PruneSSL} across various classification tasks. Each graph illustrates the average accuracy over 5 repetitions of FreeMatch during the entire learning process. The shaded region signifies the standard error for each plot. Notably, it is observed that employing \emph{PruneSSL}'s data pruning approach (depicted in orange) yields superior performance compared to training with the original unlabeled data (illustrated in black). The green line represents \emph{PruneSSL}'s performance when using an oracle instead of the pseudo-labeling function. The blue line represents training with random pruning of the unlabeled dataset.}
    \label{fig:competitve_models}
    \vspace{0.2cm}
\end{figure*}

Our approach bears a close connection to the fundamental cluster assumption of SSL \citep{chapelle2006semi}. This assumption states that in order for any semi-supervised learning framework to work, even outside the realm of deep learning, the algorithm must assume that the data have inherent cluster structure, and thus, instances falling into the same cluster have the same class label. A direct result of this assumption is that the decision boundary of SSL algorithms should avoid intersecting high-density regions of the data. Instead, it should reside within low-density regions, thereby preserving the data's cluster structure. By enhancing the separability of unlabeled data, our method guides various SSL algorithms away from solutions that might traverse high-density regions. This ultimately enhances the performance of deep SSL techniques in a broader context.

\subsection*{Related work}

In classical machine learning, training with unlabeled data could potentially lead to performance degradation \cite{chawla2005learning,yang2011effect,li2014towards}.  This phenomenon was primarily attributed to the manifold and cluster assumptions: unlabeled data is expected to be helpful only if it lies on a low-dimension manifold, and if similar classes are clustered together \cite{chapelle2006semi,singh2008unlabeled}. In contrast, the use of unlabeled data in deep learning is generally regarded as beneficial across most scenarios \citep{yang2022survey}. This divergence could be attributed to deep learning's exceptional capacity \citep{johnson2016perceptual} to map data into spaces where both the manifold and cluster assumptions hold. This paper demonstrates the mutual benefits that can be harnessed between classical machine learning insights and deep learning. While unlabeled data generally helps deep learning, forcing it to better uphold the cluster assumption can further help the performance of deep models.

Our study diverges from previous art by focusing on altering the unlabeled data directly, rather than modifying the algorithms that use it. While numerous works draw inspiration from the cluster assumption, they often drive the separation boundary to reside in less dense regions of the unlabeled data via algorithmic optimizations \citep{chapelle2005semi, ruiz2010density, verma2022interpolation}. In contrast, our approach modifies the unlabeled data itself, implicitly encouraging SSL algorithms to adhere to the cluster assumption more closely. It's important to highlight that many methods assume a shared distribution between labeled and unlabeled sets \citep{ouali2020overview, oliver2018realistic, berthelot2019mixmatch, li2020dividemix}. Therefore, our method can be used in combination with these other methods, to boost their performance.

Recently, the recognition that specific instances within the unlabeled dataset can detrimentally impact the learning process has gained prominence, influencing several semi-supervised learning (SSL) strategies \citep{ren2020not}. For instance, FixMatch \citep{sohn2020fixmatch} introduced the concept of integrating an unsupervised loss relying on pseudo-labels assigned exclusively to unlabeled instances demonstrating high model confidence. This approach effectively steers the model to learn from a selective set of examples in the unlabeled set, which dynamically evolves throughout the training process. After the introduction of FixMatch, numerous competitive methods have integrated akin selective strategies \citep{DBLP:conf/nips/ZhangWHWWOS21, berthelot2021adamatch, li2021comatch, zheng2022simmatch, fan2023revisiting, jiang2023reliamatch}. 

The primary distinction between the methods discussed above and our approach lies in the utilization of examples from the unlabeled set. In the reviewed methods, all unlabeled examples have the potential to be employed, contingent upon the confidence of the trained model. In contrast, since our approach does not interfere with the optimization process of SSL, examples that are pruned become unavailable to the model irrespective of the model's confidence in them.

Another line of relevant works draws inspiration from adversarial attacks. In the context of attacking semi-supervised methods, \citet{carlini2021poisoning, shejwalkar2022perils} demonstrated that introducing unlabeled examples that impair data separability can negatively impact the performance of several SSL algorithms. This corroborates our findings, underscoring the advantageous nature of removing analogous examples from the unlabeled set.

\section{Method}
\label{sec:method}
We consider a $K$-class classification scenario within a semi-supervised learning (SSL) framework. Here, $\mathcal{X}$ is the set of all possible examples, and $\mathcal{Y}$ is their corresponding labels. A semi-supervised learner denoted as $f:\mathcal{X}\rightarrow\mathcal{Y}$ is trained using a set of labeled examples $\mathcal{L}\subseteq\mathcal{X}\times\mathcal{Y}$ and unlabeled set of examples $\mathcal{U}\subseteq\mathcal{X}$. As labeled examples are often more expensive to obtain than unlabeled examples, traditionally the labeled pool $\mathcal{L}$ is significantly smaller than the unlabeled pool $\mathcal{U}$. The empirical generalization error of $f$ is its performance on a labeled test set distinct from both $\mathcal{L}$ and $\mathcal{U}$.

In many SSL scenarios, the labeled pool, unlabeled pool, and the test set all originate from the same underlying data distribution. In our experimental setup, we alter the distribution from which $\mathcal{U}$ is sampled. We observe that drawing $\mathcal{U}$ from the same distribution as $\mathcal{L}$ may not be ideal for learning. Specifically, our findings indicate that increasing the separability of $\mathcal{U}$, achieved by pruning hard examples from it, significantly improves the learning of $f$.

Let $\mathcal{U}'\subseteq\mathcal{U}$ denote a subset of the unlabeled pool $\mathcal{U}$, obtained by applying some pruning method. In our experimental setup, we explore three distinct unsupervised pool types: (i) $\mathcal{U}'_{rand}$ is obtained by uniformly and randomly removing examples from $\mathcal{U}$ while preserving its original distribution. (ii) $\mathcal{U}'_{prune}$ is obtained by removing points as suggested by \emph{PruneSSL}, without relying on any labels of $\mathcal{U}$ to guide it. (iii) $\mathcal{U}'_{oracle}$ is similar to $\mathcal{U}'_{prune}$, but here the removal of points by \emph{PruneSSL} is guided by the true labels of $\mathcal{U}$, rather than its using an inferred pseudo-labeling function.

\begin{figure*}[h!]
    \begin{subfigure}{1\textwidth}
      \centering
      \includegraphics[width=1\linewidth]{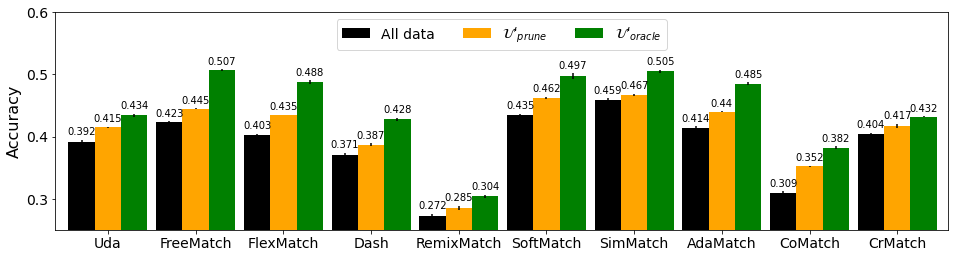}
      % \caption{$|\mathcal{U}'| = 2000$}
      % \label{subfig:clustering_matters_regular}
    \end{subfigure}
    \caption{Different SSL algorithms trained with \emph{PruneSSL}. Each group of bars depicts the mean final accuracy of $3$ WRN networks, trained with different SSL algorithms, on CIFAR-100, with $300$ labeled examples (3 per class). \emph{PruneSSL} improves all the SSL methods, despite using a smaller unlabeled dataset.}
    \label{fig:different_ssl_algorithms}
    \vspace{0.2cm}
\end{figure*}

When constructing $\mathcal{U}'_{oracle}$ and $\mathcal{U}'_{prune}$, our objective is to generate distributions that exhibit enhanced separability as compared to the original unlabeled pool $\mathcal{U}$. To achieve this, we outline a 4-step general protocol for the creation of these unlabeled datasets:
\begin{enumerate}
    \item Conduct a deep-representation task on $\mathcal{U}$, resulting in an embedding space in which the problem is more linearly separated than in the original pixel space. 
    \item Obtain labels for the unsupervised pool $\mathcal{U}$, using the real labels for $\mathcal{U}'_{oracle}$ and pseudo-labels for $\mathcal{U}'_{prune}$.
    \item Train a simple classifier using the representation from step (1) and the labels obtained in step (2). 
    \item Use the classifier's confidence to determine which examples are hardest to classify, and prune them from the unlabeled pool. 
\end{enumerate}
For a detailed algorithmic representation, please refer to Alg.~\ref{alg:our_method}. Additionally, an illustrative visualization of this process can be found in  Fig.~\ref{fig:tsne_visualization}.

\begin{algorithm}[H]
    \footnotesize
    \centering
    \caption{PruneSSL}
    \label{alg:our_method}
    \begin{algorithmic}[1]
    \STATE {\bfseries Input:} Unlabeled set $\mathcal{U}$, $n=\#$ examples in final unlabeled set
    \STATE {\bfseries Output:} $\mathcal{U}'_{prune}\subseteq \mathcal{U}$
    \STATE embedding$\_$X $\leftarrow$ self$\_$representation$\_$task$(\mathcal{U})$
    \STATE pseudo$\_$labels $\leftarrow$ pseudo$\_$labeling$\_$task$(\mathcal{U})$
    \STATE clf $\leftarrow$ simple$\_$classifier.fit$($embedding, pseudo$\_$labels$)$
    \STATE $score$ $\leftarrow$ clf.confidence(embedding)
    \STATE $\mathcal{U}'_{prune}$ $\leftarrow$ $n$ examples from $\mathcal{U}$ with highest $score$
    \STATE \textbf{return} $\mathcal{U}'_{prune}$ 
    \end{algorithmic}
\end{algorithm}

In our work, we explored various alternatives for each step outlined above, yielding consistent qualitative outcomes, as elaborated in the subsequent section. Concerning embedding methods, we employed self-representation tasks like SimCLR \citep{chen2020simple} and transfer learning based on the Inception network \citep{szegedy2015going} trained on ImageNet \citep{deng2009imagenet}. In terms of the simple classifier, we evaluated choices including a linear SVM,  SVM with an RBF kernel, or a small fully connected deep network. The SVM's confidence score was calculated based on the examples' distance from the separating hyperplane, while for the neural network, confidence was determined by the logit corresponding to the example's label. These diverse approaches consistently produced the same qualitative results, as shown in the following section.

\subsection{Technical details}

In the experiments outlined in Section~\ref{sec:results}, our evaluation considered a range of SSL algorithms. To ensure a fair comparison, we adopt the SSL evaluation environment crafted by \citep{usb2022}. This repository offers a diverse selection of SSL methods, all evaluated on the same underlying architectures and datasets. We consider the following recent SSL methods: 
Dash \citep{xu2021dash}, FlexMatch \citep{DBLP:conf/nips/ZhangWHWWOS21}, FreeMatch \citep{DBLP:conf/iclr/Wang0HHFW0SSRS023}, RemixMatch \citep{berthelot2019remixmatch}, SoftMatch \citep{DBLP:journals/corr/abs-2301-10921}, Uda \citep{xie2020unsupervised}, SimMatch \citep{zheng2022simmatch}, AdaMatch \citep{berthelot2021adamatch}, CoMatch \citep{li2021comatch} and CrMatch \citep{fan2023revisiting}. The specific architectures and hyper-parameters used for each method are detailed below.

\subsubsection{Datasets}
\label{sec:datasets}
In the experiments below, we considered 4 datasets: CIFAR-10, CIFAR-100, STL-10 \citep{coates2011analysis}, and a binary subset of CIFAR-10. When using STL-10, we omitted the unlabeled split due to its inclusion of out-of-distribution examples. The binary subset of CIFAR-10 contained the examples from CIFAR-10 that belonged to the cats and the airplanes classes. This dataset was used due to its simplicity -- while still containing real images, SSL algorithms could learn with a small labeled set, even when a small neural architecture was used, drastically reducing its computational cost. This was needed, especially in the ablation study, as state-of-the-art SSL methods are often computationally demanding.

\subsubsection{Architectures and hyper-parameters}
When training SSL methods on CIFAR-10, CIFAR-100, and STL-10 datasets, we employed the Wide-ResNet-28 (WRN) architecture \citep{zagoruyko2016wide} as the underlying architecture, using 2 width factor, stochastic gradient descent optimizer, 64 batch size, 0.03 learning rate, and 0.9 momentum. We used 0.001 weight decay for CIFAR-10 and 5e-4 for CIFAR-100 and STL-10. When training SSL methods on the binary CIFAR-10, we used a small Vision-Transformer (ViT) \citep{dosovitskiy2020image}, termed \emph{small-ViT}. \emph{small-ViT} has 6 depth, 3 attention heads, 96 width, 5e-4 learning rate, 0.9 momentum, 8 batch size, and was trained using AdamW. 

When training SCAN, we used a ResNet-18 architecture with Adam optimizer, learning rate of 1e-4, 0.9 momentum, 1e-4 weight decay, and 128 batch size. When training SimCLR, we used ResNet-18 architecture, with Adam optimzier, 3e-4 learning rate, 256 batch size, and 1e-4 weight decay.

Empirically, we observed that while small-ViT achieves significantly worse performance than SOTA architectures, the qualitative results of every experiment remain the same when small-ViT is replaced by WRN. All the experiments were performed using Nvidia GeForce RTX 2080 GPUs.

\subsection{Implementation choices for PruneSSL}
Unless explicitly stated otherwise, the implementation of \emph{PruneSSL} on CIFAR-10, CIFAR-100, or STL-10, employed SimCLR as the feature space. Subsequently, pseudo-labels were derived using SCAN, followed by RBF-SVM as the simple classifier. When running \emph{PruneSSL} on the binary CIFAR-10, we used k-means instead of SCAN to obtain pseudo-labels. In all datasets, \emph{PruneSSL} pruned $40\%$ of the data, keeping in the unlabeled set $60\%$ of the examples. For CIFAR-10 and STL-10, we used 40 labels for $\mathcal{L}$ and 300 examples For CIFAR-100. In all cases, all classes had the same number of examples in the labeled pool $\mathcal{L}$.

\section{Empirical Evaluation}
\label{sec:results}

\begin{figure*}[htb!]
    \begin{subfigure}{.33\textwidth}
      \centering
      \includegraphics[width=1\linewidth]{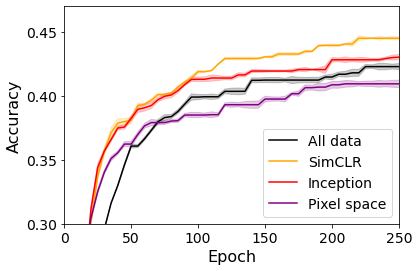}
      \caption{Different feature spaces}
      \label{fig:different_feature_spaces}
    \end{subfigure}
    \begin{subfigure}{.32\textwidth}
      \centering
      \includegraphics[width=1\linewidth]{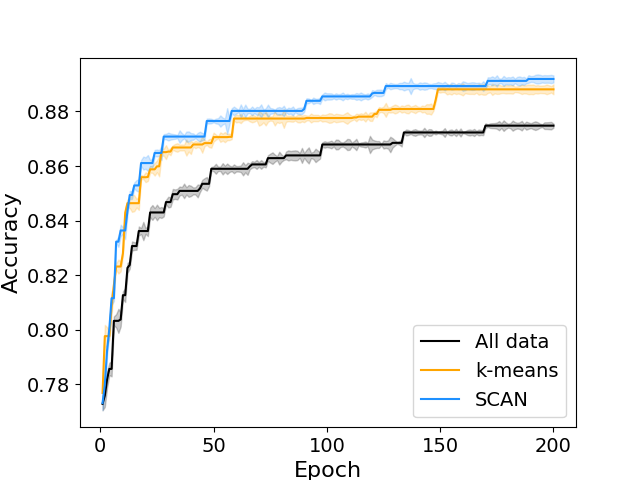}
      \caption{Different pseudo labeling}
      % \caption{$|\mathcal{U}'| = 2000$}
      \label{fig:different_pseudo}
    \end{subfigure}
    \begin{subfigure}{.33\textwidth}
      \centering
      \includegraphics[width=1\linewidth]{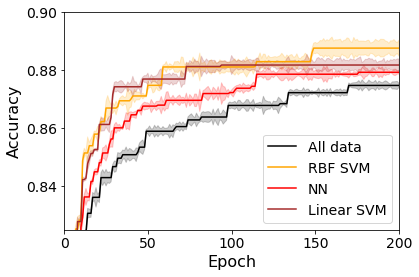}
      \caption{Different final classifiers}
      % \caption{$|\mathcal{U}'| = 2000$}
      \label{fig:different_final_classifiers}
    \end{subfigure}
    \caption{
    Evaluation of diverse choices at each stage of the \emph{PruneSSL} Algorithm. \emph{PruneSSL} begins by training a representation space, followed by a pseudo-labeling function, and culminates by training a simple classifier that effectively prunes examples with low confidence. In panel (a), we train FreeMatch on CIFAR-100, with \emph{PruneSSL} variants based on either the representation space achieved by training SimCLR or when using the penultimate layer of Inception pre-trained on ImageNet. In panel (b) we train FreeMatch on the binary CIFAR-10 dataset, with \emph{PruneSSL} variants based on either SCAN or k-means for the pseudo-labeling function. In panel (c) we train FreeMatch on the binary CIFAR-10 dataset, with \emph{PruneSSL} variants based on either linear SVM, RBF kerneled SVM, or a small fully connected neural network. Across all three panels, we plot the mean of 5 repetitions, where the shaded area denotes the standard error.  We see that whatever method is eventually used in each step, training with the data achieved by \emph{PruneSSL}, despite being significantly smaller, is better than training with the entire unlabeled set. These results demonstrate the adaptability of \emph{PruneSSL} and its effectiveness.}
    \label{fig:different_parts_of_the_algorithm}
\end{figure*}

\paragraph{PruneSSL enhances SSL algorithms on different datasets.}

% Figure~\ref{fig:competitive_models} illustrates a performance comparison of FreeMatch using distinct unlabeled data sets: $\mathcal{U}'_{prune}$, $\mathcal{U}'_{oracle}$, $\mathcal{U}'_{rand}$, and the comprehensive unlabeled set $\mathcal{U}$. This assessment is conducted across CIFAR-10, CIFAR-100, and STL-10 datasets, employing a competitive WRN architecture. Observing all datasets, it becomes apparent that while $\mathcal{U}'_{rand}$ exhibits slightly reduced performance when contrasted with the complete unlabeled dataset $\mathcal{U}$, both $\mathcal{U}'_{prune}$ and $\mathcal{U}'_{oracle}$ demonstrate markedly enhanced outcomes.

% Despite the fact that $\mathcal{U}$, $\mathcal{U}'_{rand}$, and $\mathcal{U}'_{oracle}$ maintain class balance, this equilibrium is not maintained in $\mathcal{U}'_{prune}$ due to the absence of labels during its construction. Remarkably, even with this class imbalance, $\mathcal{U}'_{prune}$ still surpasses the performance of $\mathcal{U}'_{rand}$. Importantly, $\mathcal{U}'_{oracle}$ outperforms $\mathcal{U}'_{prune}$, underscoring the potential for augmenting PruneSSL's efficacy in the future by refining the pseudo-labeling mechanism.

Fig.~\ref{fig:competitve_models} illustrates a performance comparison of FreeMatch using different sets of unlabeled data: $\mathcal{U}'_{prune}$, $\mathcal{U}'_{oracle}$, $\mathcal{U}'_{rand}$, and the complete unlabeled set $\mathcal{U}$. This evaluation is conducted on CIFAR-10, CIFAR-100, and STL-10 datasets, utilizing a competitive WRN architecture. Across all datasets, it is evident that while $\mathcal{U}'_{rand}$ exhibits slightly inferior performance as compared to the entire unlabeled dataset $\mathcal{U}$, both $\mathcal{U}'_{prune}$ and $\mathcal{U}'_{oracle}$ showcase significantly improved results. 

Despite the fact that $\mathcal{U}$, $\mathcal{U}'_{rand}$, and $\mathcal{U}'_{oracle}$ maintain class balance, this balance is not maintained in $\mathcal{U}'_{prune}$ due to the absence of labels during its construction. Remarkably, even with this class imbalance, $\mathcal{U}'_{prune}$ still surpasses the performance of $\mathcal{U}'_{rand}$. As expected, $\mathcal{U}'_{oracle}$ outperforms $\mathcal{U}'_{prune}$, indicating that potential improvements in the pseudo-labeling function hold promise for advancing \emph{PruneSSL}'s performance in the future.

\paragraph{PruneSSL improves different SSL methods.}

The results depicted in Fig.~\ref{fig:competitve_models} extend beyond the FreeMatch algorithm. In the experiments depicted in Fig.~\ref{fig:different_ssl_algorithms}, we focus on CIFAR-100 and perform the same experiment as Fig.~\ref{fig:competitve_models}, but with a wide variety of SSL algorithms. For each algorithm, we plot the mean final accuracy of $3$ repetitions using $\mathcal{U}'_{oracle}$, $\mathcal{U}'_{prune}$ and the entire unlabeled set $\mathcal{U}$. Analogous to the findings observed with FreeMatch, all the considered SSL algorithms show the same qualitative results -- training with \emph{PruneSSL} significantly improves the learning, despite its using fewer unlabeled datapoints. Since the methods are vastly different from each other, such results indicate that potential improvements in the SSL algorithms in the future can also advance the performance of \emph{PruneSSL}.

\subsection{Ablation study}
\label{sec:ablation_study}
In this section, we check the significance of the specific selections done by \emph{PruneSSL}.  Our investigation reveals a notable degree of robustness in the results. Diverse choices exhibit analogous qualitative trends and provide similar benefits.

\paragraph{Different deep representation feature spaces} can achieve comparable qualitative results. In Fig.~\ref{fig:different_feature_spaces}, we train FreeMatch on CIFAR-100 and contrast multiple variations of \emph{PruneSSL} against training with the complete unlabeled dataset. These \emph{PruneSSL} variations diverge in the embedding part of the algorithm. We explore the following deep-representation spaces: \begin{inparaenum}[(i)] \item SimCLR (as in Fig.~\ref{fig:competitve_models}); \item the penultimate layer of an Inception network, pre-trained on ImageNet; and \item the pixel-space of the images, without any embedding\end{inparaenum}.

% \begin{figure}[htb!]
%     \begin{subfigure}{.45\textwidth}
%       \centering
%       \includegraphics[width=1\linewidth]{graphs/different_feature_spaces/inception_vs_simclr_pixel_cifar100.png}
%       % \caption{$|\mathcal{U}'| = 2000$}
%       % \label{subfig:clustering_matters_regular}
%     \end{subfigure}
%     \caption{
%     Comparing the training FreeMatch alongside different variants of \emph{PruneSSL}, based on distinct representation spaces. Each plot represents the mean accuracy of $5$ WRN networks trained with FreeMatch on CIFAR-100 across the entire learning process. The shaded area represents the standard error of each plot. Noteworthy is the fact that despite the disparities inherent in the representation spaces, \emph{PruneSSL} consistently improves the overall performance.}
%     \label{fig:different_feature_spaces}
% \end{figure}

The deep representation spaces being compared have different characteristics: SimCLR is a contrastive-learning-based self-representation task, which is optimized specifically to get a meaningful representation space. On the other hand, the penultimate layer of Inception is initially tailored for a stand-alone classification task, later employed here in a manner analogous to transfer learning. Despite these differences, both representation spaces outperform training with the entire unlabeled set $\mathcal{U}$ by a significant margin. This result demonstrates the robustness of \emph{PruneSSL} in incorporating a wide range of representation spaces.

It's worth noting that while various deep representation spaces can be effective, not all representations are inherently suitable for \emph{PruneSSL}. As depicted in Fig.~\ref{fig:different_feature_spaces}, we see that employing the pixel space as a representation yields worse results than training with the entire unlabeled dataset.

\paragraph{Different pseudo labeling methods} can achieve similar qualitative results. Illustrated in Fig.~\ref{fig:different_pseudo}, we conduct FreeMatch training on the binary CIFAR-10 and contrast two variants of \emph{PruneSSL} against training with the complete unlabeled dataset. These \emph{PruneSSL} variants diverge in the pseudo-labeling method. We consider using either SCAN or k-means with $k=2$ on the feature space. Both variants outperform training with the entire unlabeled set $\mathcal{U}$ by a significant margin, suggesting that \emph{PruneSSL} can incorporate different pseudo-labeling techniques.

% \begin{figure}[htb!]
%     \begin{subfigure}{.45\textwidth}
%       \centering
%       \includegraphics[width=1\linewidth]{graphs/different_pseudo/cats_vs_airplanes.png}
%       % \caption{$|\mathcal{U}'| = 2000$}
%       % \label{subfig:clustering_matters_regular}
%     \end{subfigure}
%     \caption{
%     Comparing the training FreeMatch alongside different variants of \emph{PruneSSL}, based on distinct pseudo-labeling functions. Each plot represents the mean accuracy of $5$ networks trained with FreeMatch on the binary CIFAR-10 dataset. The shaded area represents the standard error of each plot. Noteworthy is the fact that despite the disparities inherent in the pseudo-labeling functions, \emph{PruneSSL} consistently improves the overall performance.}
%     \label{fig:different_pseudo}
% \end{figure}

% PruneSSL hinges on a simplistic classifier to determine which datapoints should be omitted. In all the aforementioned experiments, we employed an SVM with an RBF kernel due to its commendable performance and simplicity. Fig.~\ref{fig:different_final_classifiers} explores PruneSSL variants integrated with diverse classifier types, each exhibiting the same consistent trend. We conducted FreeMatch training on a binary subset of CIFAR-10, applying PruneSSL with linear SVM, RBF-kernel SVM, and a compact fully connected neural network. Notably, while the RBF-kernel SVM yields optimal performance, all three classifiers surpass the efficacy of training with the complete unlabeled set $\mathcal{U}$, signifying PruneSSL's compatibility with a diverse array of classifiers.

\paragraph{Changing the classifier.}
\emph{PruneSSL} relies on the confidence of a simple classifier to determine which examples should be pruned. In all the aforementioned experiments, we employed an SVM with an RBF kernel due to its simplicity and relatively good performance. Fig.~\ref{fig:different_final_classifiers} explores \emph{PruneSSL} variants integrated with diverse classifiers, each exhibiting the same qualitative behavior.  We conducted FreeMatch training on a binary subset of CIFAR-10, applying \emph{PruneSSL} with linear SVM, RBF-kernel SVM, and a small fully connected neural network. Notably, while the RBF-kernel SVM yields optimal performance, all three classifiers outperform training with the entire unlabeled set $\mathcal{U}$, suggesting that \emph{PruneSSL} is compatible with a diverse array of classifiers.

% \begin{figure}[htb!]
%     \begin{subfigure}{.45\textwidth}
%       \centering
%       \includegraphics[width=1\linewidth]{graphs/different_classifiers/svm_vs_nn_vs_linear_cats_airplanes.png}
%       % \caption{$|\mathcal{U}'| = 2000$}
%       % \label{subfig:clustering_matters_regular}
%     \end{subfigure}
%     \caption{
%     Comparing the training FreeMatch alongside different variants of \emph{PruneSSL}, based on different classifiers for confidence. Each plot represents the mean accuracy of $5$ networks trained with FreeMatch on the binary CIFAR-10 dataset. The shaded area represents the standard error of each plot. Noteworthy is the fact that despite the disparities inherent in the final classifiers that were used, \emph{PruneSSL} consistently improves the overall performance.}
%     \label{fig:different_final_classifiers}
% \end{figure}

\paragraph{Manipulating the size of the labeled set $\mathcal{L}$.}

\begin{figure}[htb!]
    \begin{subfigure}{.45\textwidth}
      \centering
      \includegraphics[width=1\linewidth]{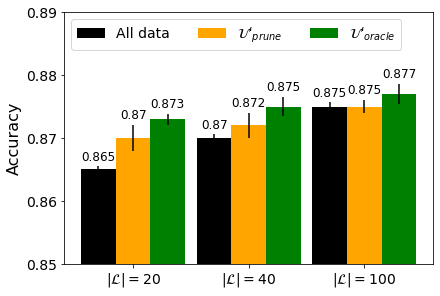}
      % \caption{$|\mathcal{U}'| = 2000$}
      % \label{subfig:clustering_matters_regular}
    \end{subfigure}
    \caption{
    Performance analysis of \emph{PruneSSL} across various sizes of the labeled set. The mean final accuracy of $5$ networks trained with FreeMatch on the binary CIFAR-10 dataset is plotted in bar groups. Standard error bars are included. Notably, \emph{PruneSSL} demonstrates more significant improvements with smaller labeled datasets.
    }
    \label{fig:different_label_set_sizes}
\end{figure}

The benefit of pruning examples from $\mathcal{U}$ becomes more pronounced when the size of the labeled set $\mathcal{L}$ is small. In Fig.~\ref{fig:different_label_set_sizes}, we present the outcomes of FreeMatch training on the binary CIFAR-10, employing varying sizes for the labeled set $\mathcal{L}$. Evidently, as the size of $\mathcal{L}$ increases, the advantage of \emph{PruneSSL} decreases. This result implies that pruning examples from $\mathcal{U}$ is more effective when the labeled pool is smaller.

\begin{figure}[htb!]
    \begin{subfigure}{.45\textwidth}
      \centering
      \includegraphics[width=1\linewidth]{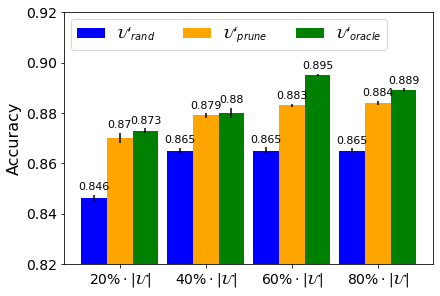}
      % \caption{$|\mathcal{U}'| = 2000$}
      % \label{subfig:clustering_matters_regular}
    \end{subfigure}
    \caption{
    Performance analysis of \emph{PruneSSL} across various sizes of the unlabeled set. The mean final accuracy of $5$ networks trained with FreeMatch on the binary CIFAR-10 dataset is plotted in bar groups. Standard error bars are included. Remarkably, while \emph{PruneSSL} achieves peak performance at $60\%$ data retention, diverse pruning levels can consistently improve the performance of the SSL algorithm.}
    \label{fig:different_unlabel_set_sizes}
\end{figure}

\paragraph{Manipulating the size of the unlabeled set $\mathcal{U}$.}
While an optimal number of examples for pruning exists, a broad spectrum of pruned example quantities can enhance SSL algorithms. Highlighted in Fig.~\ref{fig:different_unlabel_set_sizes}, we showcase the outcomes of FreeMatch training on the binary CIFAR-10, incorporating varying sizes for the unlabeled set $\mathcal{U}$. Strikingly, across all selections, a consistent qualitative trend emerges -- $\mathcal{U}'_{oracle}$ surpasses $\mathcal{U}'_{prune}$, which, in turn, outperforms $\mathcal{U}'_{rand}$. Notably, the best results are achieved when retaining $60\%$ of the data. Nevertheless, we note that a large array of other pruning values also enhance the SSL algorithm performance.

\paragraph{Adding the pruned examples back to the data.}
\label{sec:kmeans}
A possible explanation for \emph{PruneSSL}'s success might be drawn from the realm of curriculum learning \citep{bengio2009curriculum,hacohen2019power}. In curriculum learning, a learner is progressively trained on tasks of increasing complexity. The underlying concept is that mastering simpler tasks facilitates the acquisition of more complex ones. This parallel might hold here: pruning data could arguably make the problem simpler, given the enhanced separability of the unlabeled data. Once mastery over the easier version of the task is achieved, reintroducing the entire unlabeled dataset might prove advantageous.

\begin{figure}[htb!]
    \begin{subfigure}{.45\textwidth}
      \centering
      \includegraphics[width=1\linewidth]{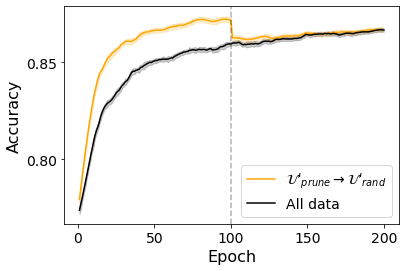}
      % \caption{$|\mathcal{U}'| = 2000$}
      % \label{subfig:clustering_matters_regular}
    \end{subfigure}
    \caption{
    Impact of reintroducing the pruned examples back to the unlabeled set. We plot the mean accuracy of 5 networks, trained with FreeMatch on the binary CIFAR-10 dataset. The black line shows a $200$-epoch training on full unlabeled dataset, while the orange line shows \emph{PruneSSL} ($100$ epochs) followed by the full unlabeled dataset ($100$ epochs). The dashed vertical line marks the reintroduction of pruned examples. The shaded area represents the standard error. We see that reintroducing the pruned examples harms the performance, suggesting that the pruned examples indeed have a negative impact on SSL algorithms.}
    \label{fig:curriculum}
\end{figure}

We conducted the following experiment: we pruned the binary CIFAR-10 data according to \emph{PruneSSL} and subjected it to $100$ epochs of FreeMatch training. Subsequently, the pruned examples were reintegrated into the unlabeled dataset, followed by an additional 100 epochs of training. %Once trained, we added back the pruned examples to the unlabeled dataset and trained for $100$ additional epochs. 
We then compared these results to those obtained when training the model over 200 epochs using the complete unlabeled dataset, without any pruning. The results are shown in Fig.~\ref{fig:curriculum}.

Inspecting Fig.~\ref{fig:curriculum}, we observe that the reintroduction of pruned examples after several iterations does not yield learning benefits. Instead, upon reintegrating the pruned examples, performance drops, converging to the same levels as training without any pruning. These results imply that the pruned examples indeed have a negative impact on learning, indicative of more than just increased difficulty. Conceivably, these examples lead SSL algorithms to choose suboptimal separation boundaries.

\paragraph{Discriminability vs coverage}
\emph{PruneSSL} is designed to amplify discriminability within unlabeled data, a tactic that inadvertently leads to a more constrained coverage of the unlabeled dataset, as certain parts of the distribution are pruned completely. In this section, we compare \emph{PruneSSL} and a pruning technique that focuses on covering all parts of the unlabeled distribution.

Coverage emerges as a pivotal concept in the active learning domain \citep{ren2021survey}. In active learning, the learner has access to a small set of labeled data and a large set of unlabeled examples. The goal is to pick examples from the unlabeled pool to be annotated so that the resulting labeled set would be optimal for the learner's performance. Several active learning approaches underscore the merit of annotating examples that cover the entire unlabeled dataset for the learners \citep{sener2017active, kirsch2019batchbald, mahmood2021low}.

To cultivate a coverage-based pruning, we draw inspiration from the work of \citet{sener2017active}, replacing \emph{PruneSSL}'s pseudo-labeling with k-means clustering with a large $k$ to cover the data. This approach prunes instances situated farthest from the cluster centroids, allowing the advantage of keeping examples from diverse parts of the distribution, while also preserving representativeness, grounded in their proximity to respective centroids.

\begin{figure}[htb!]
    \begin{subfigure}{.45\textwidth}
      \centering
      \includegraphics[width=1\linewidth]{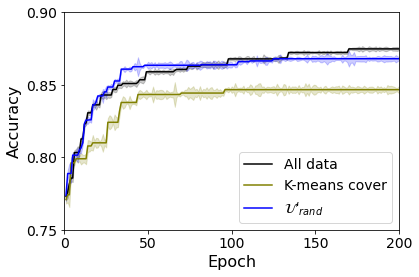}
      % \caption{$|\mathcal{U}'| = 2000$}
      % \label{subfig:clustering_matters_regular}
    \end{subfigure}
    \vspace{-.4cm}
    \caption{
    Coverage-focused pruning evaluation versus discriminability. We plot the mean accuracy of 5 networks, trained with FreeMatch on the binary CIFAR-10 dataset. The shaded area represents the standard error. The green line illustrates coverage-driven pruning using k-means centroids (see Section~\ref{sec:kmeans}). Notably, compared to random pruning (blue), k-means pruning exhibits inferior outcomes. This discrepancy may stem from the coverage goal to represent all the areas in the distribution, disproportionately representing obstructive areas, thus negatively affecting data discriminability. Results are shown for $k=50$; diverse $k$ values yield similar trends.}
    \label{fig:cover_vs_sep}
\end{figure}

In Fig.~\ref{fig:cover_vs_sep}, we plot the results of applying this coverage-oriented pruning technique to the binary CIFAR-10 dataset. A comparison between the removal of instances selected via the aforementioned method and random example removal reveals diminished performance. This outcome aligns with our initial hypothesis, as increasing the coverage of the unlabeled data decreases its discriminability.

\section{Summary and Discussion}

In this paper, we propose a way to improve the unlabeled dataset used in SSL algorithms by making it more separable. Our practical approach called \emph{PruneSSL}, focuses on pruning examples that hinder the separability of the unlabeled data, thus highlighting the inherent cluster structure of the data. The paper presents a comprehensive empirical investigation, demonstrating that this pruning technique notably enhances the performance of various competitive SSL algorithms across a diverse range of image classification tasks.

The structure of \emph{PruneSSL} involves a sequence of steps: a self-representation task followed by pseudo-labeling, and finally, a simple classifier. Through an in-depth analysis presented in the paper, we observe that each of these individual components are adaptable to a different method, suggesting that \emph{PruneSSL} could accommodate future improvements in each respective field. Testing the limits of its effectiveness, we find that \emph{PruneSSL} yields better results when the labeled dataset is smaller and the data itself is more challenging. 

\section*{Acknowledgement}
This work was supported by grants from the Israeli Council of Higher Education and the Gatsby Charitable Foundations.

\bibliography{aaai23}

% \clearpage
% \appendix
% \section*{Appendix}
% \section{More results}
% \label{app:more_results}
% This is an appendix
% \begin{figure}[htb!]
%     \begin{subfigure}{.33\textwidth}
%       \centering
%       \includegraphics[width=1\linewidth]{graphs/uda/cifar10_cats_airplanes_uda_20_labels.png}
%       \caption{$|\mathcal{L}| = 20$}
%       % \label{subfig:clustering_matters_regular}
%     \end{subfigure}
%     \begin{subfigure}{.33\textwidth}
%       \centering
%       \includegraphics[width=1\linewidth]{graphs/uda/cifar10_cats_airplanes_uda_40_labels.png}
%       \caption{$|\mathcal{L}| = 40$}
%       % \label{subfig:clustering_matters_inverse}
%     \end{subfigure}
%     \begin{subfigure}{.33\textwidth}
%       \centering
%       \includegraphics[width=1\linewidth]{graphs/uda/cifar10_cats_airplanes_uda_100_labels.png}
%       \caption{$|\mathcal{L}| = 100$}
%       % \label{subfig:clustering_matters_nocluster}
%     \end{subfigure}
%     \caption{Some text}
%     % \label{fig:clustering_matters}
% \end{figure}
% \section{Hyper-parameters}
% \label{app:hyper_params}

\end{document}